# A comparative study of deep learning and ensemble learning to extend the horizon of traffic forecasting

*Xiao Zheng[a,*], Saeed Asadi Bagloee[a], Majid Sarvi[a]*

*[a] Department of Infrastructure Engineering, The University of Melbourne, Parkville, VIC 3010, Australia*

**Abstract**: Traffic forecasting is vital for Intelligent Transportation Systems, for which Machine Learning (ML) methods have been extensively explored to develop data-driven Artificial Intelligence (AI) solutions. Recent research focuses on modelling spatial-temporal correlations for short-term traffic prediction, leaving the favourable long-term forecasting a challenging and open issue. This paper presents a comparative study on large-scale real-world signalized arterials and freeway traffic flow datasets, aiming to evaluate promising ML methods in the context of large forecasting horizons up to 30 days. Focusing on modelling capacity for temporal dynamics, we develop one ensemble ML method, eXtreme Gradient Boosting (XGBoost), and a range of Deep Learning (DL) methods, including Recurrent Neural Network (RNN)-based methods and the state-of-the-art Transformer-based method. Time embedding is leveraged to enhance their understanding of seasonality and event factors. Experimental results highlight that while the attention mechanism/Transformer framework is effective for capturing long-range dependencies in sequential data, as the forecasting horizon extends, the key to effective traffic forecasting gradually shifts from temporal dependency capturing to periodicity modelling. Time embedding is particularly effective in this context, helping naive RNN outperform Informer by 31.1% for 30-day-ahead forecasting. Meanwhile, as an efficient and robust model, XGBoost, while learning solely from time features, performs competitively with DL methods. Moreover, we investigate the impacts of various factors like input sequence length, holiday traffic, data granularity, and training data size. The findings offer valuable insights and serve as a reference for future long-term traffic forecasting research and the improvement of AI's corresponding learning capabilities.

# 1 Introduction

Traffic forecasting aims to predict future traffic conditions in a given area based on historical observations. It is a vital component of establishing an efficient Intelligent Transportation Systems (ITS) environment, with many practical applications such as transportation resource allocation and travel planning (Lieu, 2000). A predominant change in ITS recently is that extensive data can be collected from various sources, which prompts the growth of data-driven methods for traffic forecasting. In contrast to knowledge-driven methods (traditional abstract models) employing analytical or simulation models (Cascetta, 2013), data-driven methods learn traffic dynamics directly from traffic data, resulting in increased accuracy and robustness (Van Lint & Van Hinsbergen, 2012). Early attempts with data-driven methods apply statistical methods to build functional relationships, while recent research focuses on Machine Learning (ML) methods capable of capturing dynamic and complex traffic patterns (J. Liu, Wu, Qiao, & Li, 2021).

In the context of data-driven methods, traffic forecasting has focused on short-term (up to 1 hour) prediction (Lana, Del Ser, Velez, & Vlahogianni, 2018; Tedjopurnomo, Bao, Zheng, Choudhury, & Qin, 2020). This is more pronounced in recent booming research on traffic forecasting considering spatial-temporal dependency with advanced ML methods such as DCRNN (Yaguang Li, Yu, Shahabi, & Liu, 2017), STGCN (Bing Yu, Yin, & Zhu, 2017), GMAN (C. Zheng, Fan, Wang, & Qi, 2020) and TinT (W. Zheng et al., 2023). However, there is a growing interest in forecasting a longer time into the future while maintaining certain granularity (e.g., generating predictions with hourly intervals instead of daily intervals). The extended forecasting horizon can provide traffic management agencies with more time to react to traffic conditions or to prepare for special events (Bogaerts, Masegosa, Angarita-Zapata, Onieva, & Hellinckx, 2020; Manibardo, Laña, & Del Ser, 2021). Moreover, road users can use this information to plan their trips further ahead (J. J. Yu, Markos, & Zhang, 2021). Consequently, some studies have explored long-term traffic forecasting, involving predictions spanning from a few hours (Bogaerts et al., 2020; Guo et al., 2023; Kong et al., 2024) to several days (Z. He, Chow, & Zhang, 2019; Yiqun Li, Chai, Wang, Zhang, & Qiu, 2022; Qu, Li, Li, Ma, & Wang, 2019; J. J. Yu et al., 2021).

Compared to well-studied short-term forecasting, long-term forecasting, especially when the forecasting horizon extends beyond a few days, remains an open issue in the transportation literature. Long-term forecasting presents several challenges, including the increased difficulty of capturing effective long-term dependencies (whether from historical sequences of the target sensor or from other sensors with potential spatial correlations), the necessity to understand more complex periodicity patterns (such as a mixing of weekly and holiday patterns), and the amplified impact of error accumulation when using a recursive approach (where predictions are made using previously estimated values) to produce multistep forecasts (X. Zheng, Bagloee, & Sarvi, 2024b).

Towards addressing the challenges associated with extending the horizon of traffic forecasting, this paper experimentally explores the performance of promising ML traffic forecasting methods for forecasting up to 30 days (720 steps). Specifically, we develop and evaluate one ensemble ML learning method, eXtreme Gradient Boosting (XGBoost), and a spectrum of Deep Learning (DL) methods, including multilayer Sequence to Sequence (Seq2Seq) Recurrent Neural Network (RNN) and Long Short-Term Memory Network (LSTM), as well as a state-of-the-art Transformer-based method, Informer, with the particular focus on modelling capacity in the time dimension. In contrast to the majority of existing studies that only use one type of data source (from highways or urban areas) and span less than 6 months (limiting the study of longer-term seasonal patterns and the number of training samples), our evaluation is conducted with real-world signalized arterials and freeways traffic flow datasets, all with substantial temporal coverage (two years data for model development, and one year of data for evaluation). Furthermore, we leverage time embedding to add timestamp and holiday features to these models to refine them with better seasonality and event learning ability. In addition, empirical studies are conducted to discuss different important aspects of long-term traffic forecasting, including the impact of the input sequence's length, holiday traffic (as a representation of event traffic), data granularity, and size of training data. The computation costs of the studied methods are also investigated.

As we focus on expanding the forecasting in the temporal dimension, the evaluation is performed with data collected from a single location. This allows our findings can be applied to scenarios where only one detector is deployed, or the detectors are located too far to generate any spatial dependency. Additionally, while modeling spatial correlations can add significant complexity—often making these methods computationally impractical for forecasting very long sequences—this approach typically relies on phenomena like congestion propagation. As a result, it may provide limited benefits for the extended forecasting horizons explored in this study.Nonetheless, our study can also contribute to the development of spatial-temporal modelling methods, as the studied DL methods are widely adopted as the mechanism of modelling temporal correlations in these methods (L. Bai, Yao, Li, Wang, & Wang, 2020; Cai, Wang, & Hu, 2024; L. Chen et al., 2022; Du, Li, Gong, & Horng, 2018; Guo et al., 2023; Guo, Lin, Wan, Li, & Cong, 2021; Yaguang Li et al., 2017; H. Liu et al., 2023; Méndez, Merayo, & Núñez, 2023; Qi et al., 2024; Weng et al., 2023; X. Wu et al., 2025; X. Zhang et al., 2025; C. Zheng et al., 2020; W. Zheng et al., 2023; Zhou et al., 2023).

We note that a short conference version of this paper appeared in (X. Zheng, Sarvi, & Bagloee, 2022), which was presented at the 43rd Australasian Transport Research Forum (ATRF). Our initial conference paper only discussed DL methods and did not introduce time embedding. Major changes in this manuscript include (1) the incorporation of ensemble learning methods (XGBoost) into the investigation (2) the introduction of time embedding to assist periodicity learning, which is proven to be very effective for long-term forecasting (3) the execution of experiments on more datasets covering longer time span and both signalized arterials and freeways traffic, and (4) the provision of additional analysis on impact of length of input sequence, holiday traffic, data granularity, size of training data, and a discussion about the computation costs.

## 2 Related works
### 2.1 Traffic forecasting

In the early stages of traffic forecasting research, statistical methods were commonly applied. In this category, Seasonal Auto-Regressive Integrated Moving Average (SARIMA), has been applied to 1 step (15 minutes) ahead of traffic prediction (Williams & Hoel, 2003). Another statistical approach, the Kalman Filter (KF) has also been applied to the same forecasting horizon (Emami, Sarvi, & Asadi Bagloee, 2019).

Recent efforts have shifted towards ML methods to overcome some of the shortcomings of statistical approaches, such as making assumptions about specific data properties like stationarity. Traditional ML methods like Support Vector Regression (SVR) have been applied to 1-step-ahead forecasting to generate the traffic flow of the next 3 and 15 minutes, depending on the data granularity (X. Wang, An, Tang, & Chen, 2015; C.-H. Wu, Ho, & Lee, 2004). K-Nearest Neighbours (KNN) has been used to predict traffic up to 5 steps (25 minutes) into the future (Bin Yu, Song, Guan, Yang, & Yao, 2016; L. Zhang, Liu, Yang, Wei, & Dong, 2013). An advanced improvement of ML methods is ensemble learning, which accumulates weak learners to generate better predictive performance, and this approach has been successfully applied to many fields (Sagi & Rokach, 2018; X. Zheng, Bagloee, & Sarvi, 2024a). In the realm of traffic forecasting, Random Forest (RF) has been used for up to 24 steps (1 day) ahead (Hou, Edara, & Sun, 2015). XGBoost (T. Chen & Guestrin, 2016), a scalable and efficient improvement of the Gradient Boosting Regression Tree (GBRT), has produced higher accuracy than the historical average method and LSTM for 288 steps (1 day) ahead traffic forecasting (Cao, Cen, Cen, & Ma, 2020). Another study reported that, compared with RF, XGBoost can deliver better accuracy for short-term urban traffic prediction (Alajali, Zhou, Wen, & Wang, 2018).

DL methods are a type of ML methods that employ multiple-layer architectures to extract inherent features in data. They are considered capable of identifying highly non-linear temporal dependencies and have a recognized ability to model complex traffic patterns (Tedjopurnomo et al., 2020). RNN computes an internal cell state summarising past information, enabling it to capture dependencies between different parts of the sequence, and it has been widely used for temporal forecasting problems (Salinas, Flunkert, Gasthaus, & Januschowski, 2020; Schmidt, 2019; Y. Wang et al., 2019). LSTM is a variant of RNN and is designed to enhance the model's ability to model long-range dependencies in sequences with its gated structure and memory cells. It has been widely applied to traffic forecasting, with the longest prediction span reported to be 60 minutes (Yongxue Tian & Pan, 2015; Yan Tian, Zhang, Li, Lin, & Yang, 2018; Zhao, Chen, Wu, Chen, & Liu, 2017). A Seq2Seq architecture is introduced to allow RNN-based methods to predict an output sequence of arbitrary length. A method based on Seq2Seq LSTM has shown promising results for up to one day ahead of traffic prediction (Z. Wang, Su, & Ding, 2020).

RNN and its variances force past information to be stored in vectors of fixed size, regardless of the input size, which can lead to lost information (bottleneck issue). The attention mechanism is introduced to solve this problem (Bahdanau, Cho, & Bengio, 2014). Intuitively, attention is a weighted combination of a sequence of data, where the weights are determined by the level of similarity among data points. It allows the model to search among every historical time step in the encoder, determining their importance and selecting the relevant information for predicting during the decoding procedure. Completely avoiding recurrent structure, Transformer (Vaswani et al., 2017) leverages the attention mechanism and has demonstrated superior performance in capturing long-term dependencies and interactions in sequential data (S. Li et al., 2019). Transformer consists of an encoder and a decoder, each containing multiple identical blocks. Each encoder block incorporates a multi-head self-attention (Cheng, Dong, & Lapata, 2016) module and a position-wise feed-forward network, utilizing residual connections (K. He, Zhang, Ren, & Sun, 2016) and layer normalization (Ba, Kiros, & Hinton, 2016). Decoder blocks additionally insert cross-attention modules between encoder and decoder. When applying Transformer to perform long-term prediction, a key issue is the $T$-quadratic computation and memory consumption on $T$ length inputs/outputs, which is exaggerated when extending the forecasting horizon. Informer (Zhou et al., 2023) is a successful attempt to improve the efficiency of the Transformer while maintaining or even improving accuracy.

RNN-based methods and attention mechanism/Transformer architecture are adopted as predominant backbones or components for modelling temporal dynamics in recent research on modelling spatial-temporal dependencies for traffic forecasting. Specifically, RNN-based methods have been integrated with spatial learning techniques such as Graph Neural Networks (GNNs) or Convolutional Neural Networks (CNNs) (L. Bai et al., 2020; L. Chen et al., 2022; Du et al., 2018; Yaguang Li et al., 2017; Qi et al., 2024; Weng et al., 2023). Attention mechanism/Transformer architecture has been employed to capture temporal and spatial dependencies within GNN-based methods, in which they are recognized as effective approaches for achieving long-term forecasting (Cai et al., 2024; Guo et al., 2021; H. Liu et al., 2023; X. Wu et al., 2025; X. Zhang et al., 2025; C. Zheng et al., 2020; W. Zheng et al., 2023). However, for the extended forecasting horizon examined in this study, the high complexity of modelling spatio-temporal correlations with GNN-based methods using attention mechanism/Transformer architecture can make these methods computationally prohibitive. It is also worth noting that the spatial-temporal modelling methods primarily focus on enhancing the learning of time-space relationships and are generally applied to predictions within 1-hour intervals, with limited research extending these forecasts to several days or more.

In summary, ensemble ML methods like XGBoost and DL models like RNN-based and Transformer-based models have demonstrated promising potential for long-term traffic forecasting. However, there is a need for a deeper understanding of their performance when extending forecasting horizons, particularly for larger horizons like 7 days or 30 days. Additionally, there is a lack of a generic suite for evaluating these methods under consistent and comprehensive conditions.

## 2.2 External factors utilisation

Traffic forecasting can be greatly impacted by additional factors other than the traffic time series. Timestamp information has been widely used for long-term forecasting as it can provide reliable information for the future and allow the modelling of different seasonal patterns. Comparatively, static factors (such as physical properties of the road and nearby land use) are normally assumed to remain constant throughout the entire data collection period in the existing ML traffic forecasting literature. Under this assumption, they are mainly informative for learning correlations between different sensors and can hardly contribute to effectively extending forecasting horizons (L. Chen et al., 2022). Other factors like additional time series (e.g., weather conditions, traffic time series from other data collection points) are challenging to forecast well in advance, impairing the benefits of employing them. It is also worth noting that, despite the importance of incorporating timestamp information for long-term forecasting, current studies often only consider day-of-the-week and time-of-the-day information (Guo, Lin, Feng, Song, & Wan, 2019; Guo et al., 2023; C. Zheng et al., 2020). However, a longer forecasting horizon is likely to cover a time period consisting of various types of traffic, necessitating the consideration of higher-level variations like monthly patterns and other future time-varying patterns, such as holidays and other events that can be planned or assumed for what-if analysis.

Timestamp information has been integrated into traffic forecasting through various approaches. Some work used them to guide the design of specific model frameworks (Guo et al., 2019; Guo et al., 2021; Z. He et al., 2019; J. J. Yu et al., 2021). There are also studies incorporating features extracted from timestamp information into their models by simply treating them as numeric features or embedding them with methods like one-hot encoding (L. Chen et al., 2022; C. Zheng et al., 2020), struggling with problems like capturing relationships between features (B. Wang, Shaaban, & Kim, 2021). Therefore, it would be beneficial to develop a more general solution or improved methods for feature representation learning.

# 3 Methodology

A promising ensemble ML model, XGBoost, and a spectrum of DL methods including RNN, LSTM, and the state-of-the-art Transformer-based model, Informer, are developed to analyse their performance for long-term traffic forecasting. Additionally, we utilize time embedding to incorporate timestamp and holiday features into these models, aiming to enhance their ability to capture seasonality and events. The overall methodology is summarized in Figure 1.

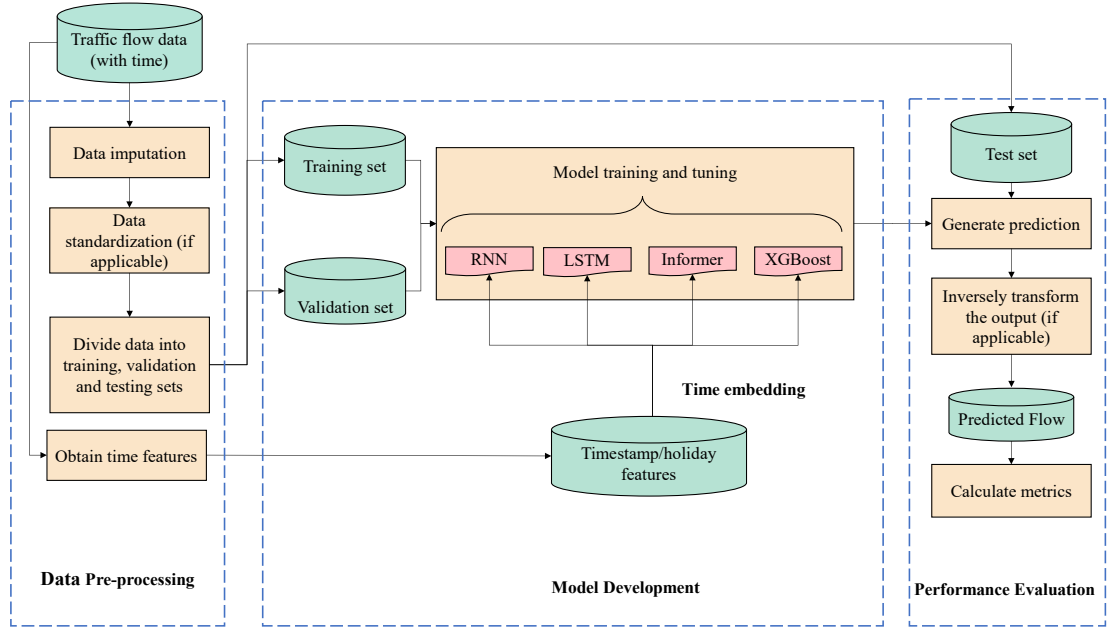

**Figure 1 Overall methodology.**

## 3.1 Problem definition

Suppose the output of interest is the flow on an isolated location and the input is historical time-series flow data collected from the same location; the traffic flow prediction problem can be stated as follows:

given the input traffic flow sequence $X = (x_1, x_2, \ldots, x_T | x_t \in \mathbb{R})$ denoting the input traffic flow of the studied road segment over $T$ time slices, predict the target traffic flow sequence $Y = (y_1, y_2, \ldots, y_{T_p} | y_t \in \mathbb{R})$ of the same road segment over the future $T_p$ slices.

## 3.2 RNN and LSTM

When predicting traffic using RNN and LSTM, a multilayer Seq2Seq framework is adopted. As shown in Figure 2, the input sequence is encoded into semantic vectors of fixed length in the encoder, which are then decoded in the decoder into the output sequence with a recursive stepwise process. Each RNN/LSTM network contains a series of layers, and each layer's output is the input into the next layer. Literature shows that the architecture of stacked layers can help capture higher-level representations of sequence data (Cui, Ke, Pu, & Wang, 2020; LeCun, Bengio, & Hinton, 2015). Note that a dropout (Srivastava, Hinton, Krizhevsky, Sutskever, & Salakhutdinov, 2014) layer is added to the outputs of each RNN/LSTM layer except the last layer, which randomly sets elements of the outputs to 0 with a designed probability $p$, and then scales the resulting outputs by a factor of $\frac{1}{1-p}$.

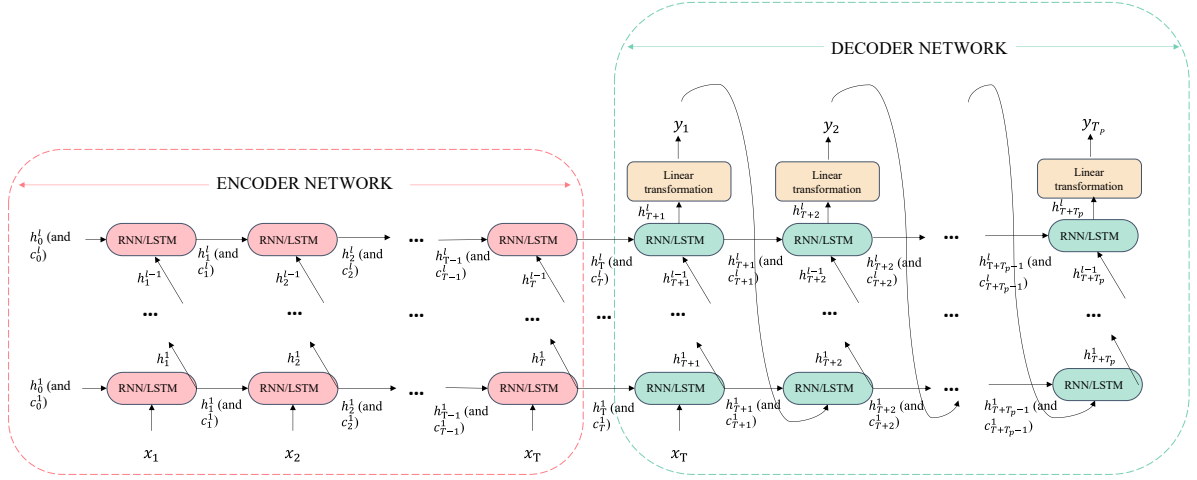

**Figure 2 Structure of a *l*-layer Seq2Seq RNN/LSTM.**

In the $m$ layer of the encoder, a sequence of $T$ length is input into the model, and each element of this sequence, $x_t^m$, contains the information of a corresponding time step. A single layer of RNN ([Schmidt, 2019](#)) can be viewed as a list of nodes, and for each node, a hidden state, $h_t^m$, is calculated with the following equation:

$$h_t^m = \tanh(W_i^m x_t^m + b_i^m + W_h^m h_{(t-1)}^m + b_h^m) \tag{1}$$

where $W_i^m$, $W_h^m$, $b_i^m$ and $b_h^m$ are trainable parameters, and $h_{(t-1)}^m$ is the hidden state from the last node, or the given initial value $h_0^m$. The generated hidden states $[h_1^m, h_2^m, \cdots, h_T^m]$ after dropout are used as the input of the subsequent layer, that is, $x_t^{m+1} = h_t^m$. Hidden states of the last step in the input are then sent into the decoder to calculate $h_{T+1}^m$ with equation (1), and the hidden state from the last layer goes through a linear transformation to obtain the prediction results, which will be used as the input to generate the next step prediction.

Compared to the multilayer Seq2Seq RNN framework, the main change in the LSTM network is the computation performed in each node. For layer $m$, each node of LSTM ([Hochreiter & Schmidhuber, 1997](#)) receives the corresponding input $x_t^m$, a hidden state $h_{t-1}^m$ and a hidden memory cell state, $c_{t-1}^m$, and maintains three gates including a forget gate $f_t^m$, a input gate $i_t^m$, and a output gate $o_t^m$. The memory cell and gates can help LSTM learn long-term dependencies by effectively storing and passing useful historical information. The calculation is performed using the following equations:

$$f_t^m = S(W_{if}^m x_t^m + b_{if}^m + W_{hf}^m h_{t-1}^m + b_{hf}^m) \tag{2}$$

$$i_t^m = S(W_{ii}^m x_t^m + b_{ii}^m + W_{hi}^m h_{t-1}^m + b_{hi}^m) \tag{3}$$

$$\tilde{c}_t^m = \tanh(W_{ic}^m x_t^m + b_{ic}^m + W_{hc}^m h_{t-1}^m + b_{hc}^m) \tag{4}$$

$$c_t^m = f_t^m \odot c_{t-1}^m + i_t^m \odot \tilde{c}_t^m \tag{5}$$

$$o_t^m = S(W_{io}^m x_t^m + b_{io}^m + W_{ho}^m h_{t-1}^m + b_{ho}^m) \tag{6}$$

$$h_t^m = o_t^m \odot \tanh(c_t^m) \tag{7}$$

where $S$ is the sigmoid function, $\odot$ is the Hadamard product ([Horn, 1990](#)), $W_{if}^m$, $W_{hf}^m$, $W_{ii}^m$, $W_{hi}^m$, $W_{ic}^m$, $W_{hc}^m$, $W_{io}^m$, $W_{ho}^m$ and $b_{if}^m$, $b_{hf}^m$, $b_{ii}^m$, $b_{hi}^m$, $b_{ic}^m$, $b_{hc}^m$, $b_{io}^m$, $b_{ho}^m$ are trainable parameters.

### 3.3 Informer

Informer ([Zhou et al., 2023](#)) is a modification of the vanilla Transformer, primarily aiming at alleviating its computation cost, which is rather valuable for long-term traffic prediction where an excessively long input sequence is needed to capture the long-range dependency. Specifically, the *ProbSparse* self-attention mechanism proposed in Informer selects and keeps only dominant dot-product pairs in attention calculation by an approximation of the *Kullback-Leibler* divergence. The resulting fundamental operation of Informer, Multi-head *ProbSparse* self-attention, is composed of $l$ mutually independent single-head *ProbSparse* self-attention layer, where the input data is mapped to the queries Q$\in \mathbb{R}^{L_Q \times d}$, keys K$\in \mathbb{R}^{L_K \times d}$ and values V$\in \mathbb{R}^{L_V \times d}$ by different linear projections, and the following calculation is conducted:

$$\mathcal{A}(Q, K, V) = \text{softmax}\left(\frac{\bar{Q}K^\top}{\sqrt{d}}\right)V \tag{8}$$

where $\bar{Q}$ only contains top $u = c \times \ln(L_Q)$ queries (where $c$ is a hyperparameter) under the sparsity measurement $M$. For $i$-th query $q_i$, $M(q_i, K)$ can be measured as:

$$M(q_i, K) = \max_j \left\{\frac{q_i k_j^\top}{\sqrt{d}}\right\} - \frac{1}{L_K} \sum_{j=1}^{L_K} \frac{q_i k_j^\top}{\sqrt{d}} \tag{9}$$

where, $k_j$ is the $j$-th key from $K$. It is proven that the calculation of $M(q_i, K)$ can only consider $U = c \times \ln(L_k)$ dot-product pairs (Zhou et al., 2020). A self-attention distilling operation is also introduced to lower the total memory usage for multiple stacking layers.

Informer also uses a generative style decoder to generate the prediction in one forward operation, which replaces the step-by-step decoding procedure in the vanilla Transformer. This design facilitates fast predicting procedures and avoids error cumulation during decoding, making it well-suited for predicting long-term sequences. At the end of all decoder layers, the results will go through layer normalization and be projected through a linear model to get the prediction.

### 3.4 XGBoost

When using XGBoost for traffic forecasting, we formulate a regression problem with it, where the inputs are time features. XGBoost is an improved version of GBRT, which uses the regression tree as the weak learner to formulate the additive model and utilizes the forward stagewise algorithm to train the model. The space of the trees can be defined as:

$$\mathcal{F} = \{f = w_{q(\mathbf{x})}\}(q: \mathbb{R}^d \to \{1, 2, \cdots, T\}, w \in \mathbb{R}^T) \tag{10}$$

where $q$ represents the formation of each tree by mapping an example to its corresponding leaf index; **x**, a $d$-dimensional vector, is the input training sample embedded with time information; $T$ is the number of leaves, and $w$ is the leaf weight. The objective function of XGBoost of iteration round $t$ is defined as:

$$\text{obj}^{(t)} = \sum_{i=1}^{n} l\left(y_i, \hat{y}_i^{(t-1)} + f_t(x_i)\right) + \Omega(f_t) + \text{constant} \tag{11}$$

where $l$ is the desired loss function, $y_i$ is the target, $\Omega(f_t)$ is a penalty term for complexity, $\hat{y}_i^{(t-1)}$ is the prediction of the $i$-th instance of round $(t-1)$, $f_t(x_i)$ is the result of the newly added tree in round $t$, $x_i$ is the $i$-th instance of training data. Apply Taylor expansion to obj$^{(t)}$ and define $g_i = \partial_{\hat{y}^{(t-1)}} l(y_i, \hat{y}^{(t-1)}), h_i = \partial^2_{\hat{y}^{(t-1)}} l(y_i, \hat{y}^{(t-1)})$ can get the approximation of obj$^{(t)}$:

$$obj^{(t)} \simeq \sum_{i=1}^{n} \left[l\left(y_i, \hat{y}_i^{(t-1)}\right) + g_i f_t(x_i) + \frac{1}{2} h_i f_t^2(x_i)\right] + \Omega(f_t) + \text{constant} \tag{12}$$

$\Omega(f_t)$ can be defined as:

$$\Omega(f_t) = \gamma T + \frac{1}{2}\lambda \sum_{j=1}^{T} w_j^2 \qquad (13)$$

where $\gamma$ and $\lambda$ are hyperparameters, $j$ is the leaf number.

Now define $I_j = \{i \mid q(\mathbf{x}_i) = j\}$ and remove the constant terms, the new objective equation can be solved as a quadratic equation of variable $w_j$:

$$\begin{aligned} Obj^{(t)} &= \sum_{i=1}^{n} \left[ g_i f_t(\mathbf{x}_i) + \frac{1}{2} h_i f_t^2(\mathbf{x}_i) \right] + \gamma T + \frac{1}{2}\lambda \sum_{j=1}^{T} w_j^2 \\ &= -\frac{1}{2} \sum_{j=1}^{T} \frac{(\sum_{i \in I} g_i)^2}{\sum_{i \in I} h_i + \lambda} + \gamma T \end{aligned} \qquad (14)$$

$Obj^{(t)}$ can be viewed as the structure score of the tree. With this newly defined structure score, XGBoost can determine the tree to be added to the ensemble model with the gain as:

$$Gain = \frac{1}{2} \left[ \frac{(\sum_{i \in I_L} g_i)^2}{\sum_{i \in I_L} h_i + \lambda} + \frac{(\sum_{i \in I_R} g_i)^2}{\sum_{i \in I_R} h_i + \lambda} - \frac{(\sum_{i \in I} g_i)^2}{\sum_{i \in I} h_i + \lambda} \right] - \gamma \qquad (15)$$

where $I_L$ and $I_R$ are instance sets of the right and left nodes splitting from the existing tree structure. Here XGBoost uses a split finding algorithm, in which the data is sorted according to feature values and then visited in this sorted order.

## 3.5 Time embedding

Aiming at enabling the prediction methods to better capture the long-term time dependency, we incorporate seven time features into the studied model. Specifically, six types of timestamp features (discrete variables) including *day of the week*, *quarter of the year*, *month of the year*, *day of the month*, *hour of the day*, and *day of the year*, and one binary holiday feature, *whether it is any public holiday,* are embedded into the models. The process of incorporating these features is termed "time embedding". It reflects the absolute location in the historical data (global information) and provides agnostic event information. Notably, although this information can be considered as external factors in addition to the input data sequence alone, it is known information about the future and can be easily accessed or directly retrieved from the collected time-series data.

For XGBoost, time embedding is implemented straightforwardly by formulating a regression problem with time features as the inputs, as described earlier. Without incorporating any lag values, the resulting method can serve as an indicator of the effect of these time features, which also has practical applications such as enabling prediction with offsets. Informer proposed a mechanism to incorporate timestamp and holiday features with learnable embedding layers and add them to the output of position embedding suggesting the order of the data points in a sequence. For Seq2Seq RNN and LSTM, we propose time embedding following (Zhou et al., 2023). In the encoder, as shown in Figure 3, after being scaled by standardization, each input flow value is transformed into a vector with a dimension of $d$ with 1-D convolutional filters which have demonstrated ability for time series feature extraction (Rizvi, 2022). $d$ filters are created during this process, and here we adopt kernel width = 3 and stride = 1. On the other hand, the time features are transferred into vector representations with the size of $d$, respectively, through entity embedding representing each time feature variable as a one-hot vector, which is then linearly projected to a vector using learnable weights. The resulting vectors are summed with the transformed flow value, and a dropout is performed to it with $p$ set as 0.1, controlling over-fitting. In the decoder, similar to the procedure shown in Figure 3, after the output of each time step is linearly projected to a scaler (and saved for later accuracy evaluation), this scaler goes through a likewise 1-D convolutional transformation and is added to the corresponding feature vectors obtained likewise. Note that time embedding can provide specifically learned and valuable information for the corresponding task; for example, it can indicate the relationship between different time features (B. Wang et al., 2021). Without further changing the magnitude, the direct summation of the 'flow vector' and time feature vectors is on the premise of the fact that the flow value is standardized, and our initial weights of embedding follow the standard normal distribution, thus the influence from each component is reasonably balanced. Besides, we do not add position embedding to RNN/LSTM as the ordering information is captured by the recurrent structure of the models.

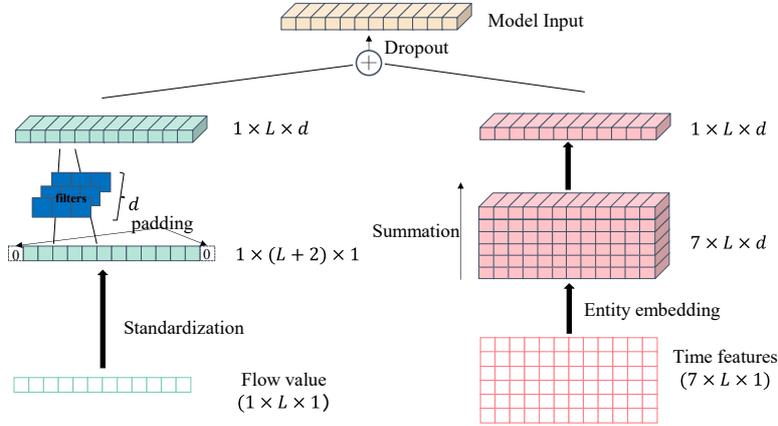

Figure 3 Time embedding for RNN and LSTM.

# 4 Performance evaluation

In this section, the proposed methods are evaluated with six large-scale real-world datasets on increasing forecasting horizons up to 30 days ahead.

## 4.1 Datasets

Six real-world traffic flow datasets are used for evaluation in this study, including:

- **Melbourne Datasets.** Three datasets are obtained from the Victorian Government (https://discover.data.vic.gov.au/dataset/traffic-signal-volume-data) (hereinafter referred to as Mel-1, Mel-2, and Mel-3, respectively). They are collected from the loop detectors installed in signalized arterials in the Melbourne urban area, whose locations are shown in Figure 4 (a). Imputation is conducted to negative, zero, and missing values, which does not take up more than 0.2% for any dataset, showing excellent fidelity. Considering the small proportion of data requiring imputation, we simply use the data from the same period last week to replace these data. The information on public holidays is collected from https://pypi.org/project/holidays/.

- **PeMS Datasets.** Three datasets and the dates of public holidays are provided by Caltrans PeMS (https://pems.dot.ca.gov/). They are observed from vehicle detector stations (VDS) located in the freeways of California, and are named as PeMS-1, PeMS-2, and PeMS-3. Their locations are shown in Figure 4 (b). Data from the same period last week is used for further imputation performed to missing values, which are at most only 0.015% of the total data for any individual dataset.

Traffic flows from all datasets are aggregated into 1-hour intervals following (Hou et al., 2015). Each of the six datasets contains three consecutive years of data starting from 1 January 2017, providing $3 \times 365 \times 24 = 26280$ data points. Data from the first 20 months (January 2017-August 2018) is used as the training set, which allows the model to capture any types of temporal patterns existing in an entire year. The validation set is the following 4 months (September-December 2018), and the trained model is tested on year 2019 data. The considerable time span of utilized data allows the models to learn from and be tested on different traffic scenarios.

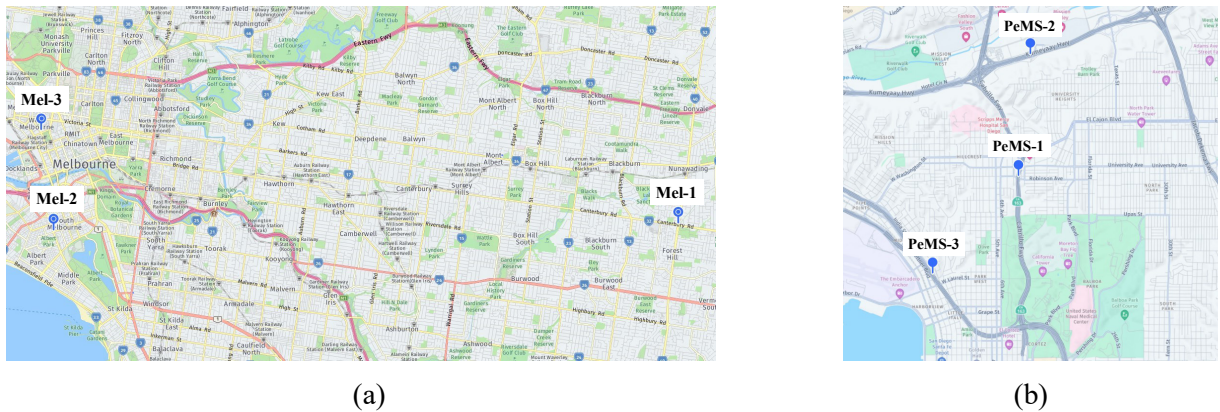

**Figure 4 Locations of Melbourne Datasets (a) and PeMS Datasets (b).**

The distribution and key statistical properties of the six datasets after imputation are calculated from the year 2017 data and illustrated in the box plot in Figure 5. Moreover, average traffic flow per day of the week and average traffic flow per month are calculated from the year 2017 data and plotted in Figure 6 and Figure 7, respectively. These visualizations reveal that the analysed datasets encompass diverse road types characterized by varying flow magnitudes, distributions, and patterns.

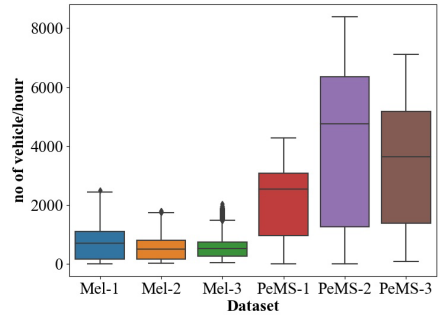

**Figure 5 Distribution of traffic datasets, calculated over year 2017.**

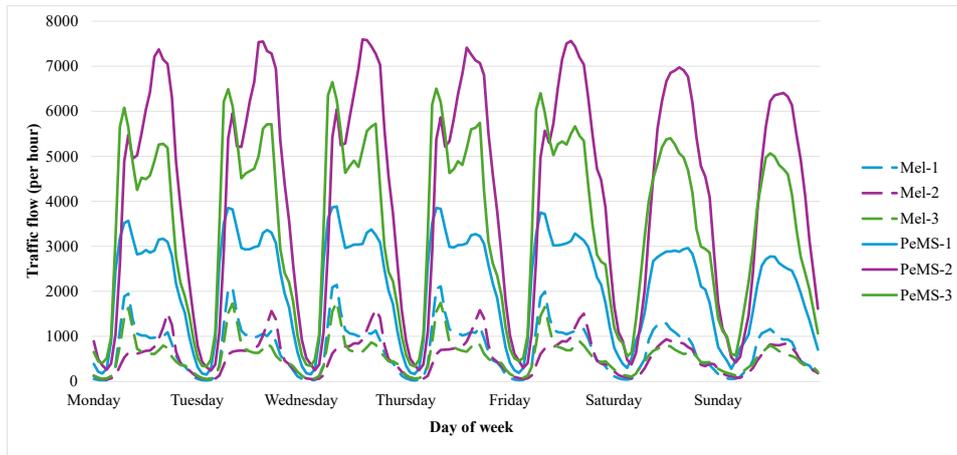

**Figure 6 Average traffic flow per each day of the week for Melbourne Datasets and PeMS Datasets (b), calculated over year 2017.**

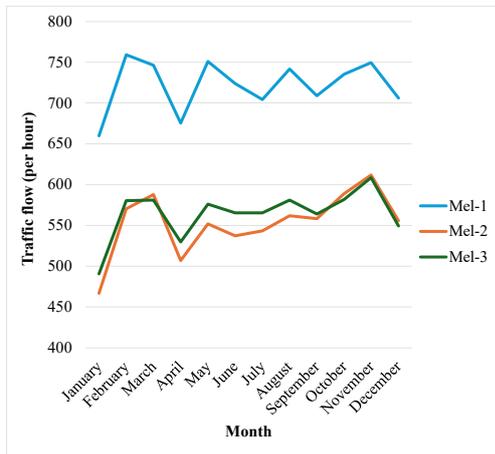
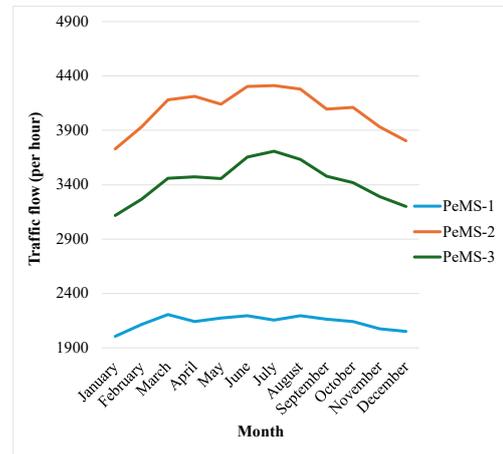

(a)  (b)

**Figure 7 Average traffic flow per month for Melbourne Datasets (a) and PeMS Datasets (b), calculated over year 2017.**

For DL methods, the data of the designed size is fed into the model with a rolling stride of 1 timestep. The flow data has been scaled by standardization with the statistics from the training set before being loaded into the models. The predicted values from the models are then inversely transformed for metric calculation. On the other hand, we fit the XGBoost one-off on the entire training dataset without performing standardization, as this method is based on decision trees, while the testing is conducted with the same 1 timestep rolling style to ensure a fair comparison. Before metrics calculation, all negative prediction results are set as 0.

## 4.2 Evaluation metrics

We use commonly employed metrics for evaluation, including Mean Absolute Error (MAE) and Root Mean Squared Error (RMSE). Inspired by (Lippi, Bertini, & Frasconi, 2013), to place greater emphasis on higher levels of traffic flow (more crucial scenarios in practical application), we also calculate Mean Absolute Percentage Error (MAPE) only for results with the corresponding ground truth exceeding 100 vehicles per hour, denoted $MAPE_{100}$.

Specifically, an additional metric used is the GEH statistics, a practical metric named after its inventor Geoffrey E. Havers (Feldman, 2012), is an empirical calculation commonly employed in the industry to gauge the goodness-of-fit of a model, defined as:

$$GEH = \sqrt{\frac{2(M-C)^2}{M+C}} \qquad (16)$$

where $M$ is the hourly traffic volume from model prediction and $C$ is the corresponding ground truth. An estimate generating a GEH index of 5.0 or less is deemed acceptable, while any GEH calculation greater than 5.0 would require attention and scrutiny. The results from the model are considered unacceptable if the GEH is above 10 (VicRoads, 2019). The use of GEH statistics will help understand if the performance is satisfying in the real-world application for the traffic prediction problem.

## 4.3 Implementation details

All DL models are implemented by PyTorch (https://pytorch.org), while XGBoost is implemented by its Python library (https://xgboost.readthedocs.io/en/stable/python/python_intro.html). The experiments are performed on one Intel(R) Xeon(R) Gold 6326 CPU @ 2.90GHz and a single NVIDIA A100 GPU (80GB RAM) with the following implementation details:

- For Informer, the architecture is set as the originally proposed model (Zhou et al., 2020) and we adopt *ProbSparse* attention (and the corresponding multi-head attention settings) for both encoders and decoders where applicable. Input length is set as 720 steps (selection of input length is discussed later in Section 4.4.2). Following (Zhou et al., 2020), Informer is trained with batch size 32, MSE loss function, and an initial learning rate of 0.0001. The initial learning rate is then reduced by multiplying $0.5^{no.of\ last\ epoch}$. Optimization is performed using the Adam optimizer (Kingma & Ba, 2015). Where applicable, early stopping is performed by monitoring the validation error, with patience = 3, which means if the validation loss increases in 3 consecutive epochs, the training is terminated, and the model with the least validation loss is saved. Otherwise, the models are trained for 100 epochs.

- For multilayer Seq2Seq RNN and LSTM, both the encoder and decoder contain 3 layers, and the size of each hidden state (and cell state for LSTM) is set as 512. For the dropout layer, $p = 0.1$. For time embedding, $d = 512$. The initial learning rate is tuned on the validation set with the range {0.01,0.001,0.0001,0.00001} and determined to be 0.0001. The same input length, batch size, loss function, learning rate decay, optimizer, and early stopping strategy used for Informer are adopted.

- For XGBoost,100 rounds of random search are performed in 2017 and 2018 data to tune the key hyperparameters. We conduct the 5-fold forward chaining validation, which only tests on the data fold with the model trained on previous data, avoiding using future patterns to predict the past. The explanation and determination of key hyperparameters tuned are summarized in Table 1. The same early stopping strategy with DL methods is performed to further control overfitting.

Table 1 Key hyperparameters tuned for XGBoost

| Hyper Parameters | Explanation | Search Range | Adopted Value | | | | | |
|---|---|---|---|---|---|---|---|---|
| | | | Mel-1 | Mel-2 | Mel-3 | PEMS-1 | PEMS-2 | PEMS-3 |
| min_child_weight | the required minimum value for the sum of leaf node weight | [1, 5] | 4 | 1 | 3 | 3 | 4 | 4 |
| gamma | required minimum loss reduction associated with further partitioning of a leaf node | (0, 0.5) | 0.378 | 0.413 | 0.445 | 0.362 | 0.155 | 0.019 |
| subsample | the proportion of used training samples to all samples for every boosting round | (0.5, 1) | 0.696 | 0.706 | 0.796 | 0.942 | 0.964 | 0.787 |
| colsample_bytree | fraction of features to be subsampled when constructing a tree | (0.5, 1) | 0.746 | 0.771 | 0.638 | 0.953 | 0.854 | 0.538 |
| max_depth | assigned maximum depth of trees | [3, 20] | 8 | 4 | 7 | 7 | 6 | 6 |
| learning_rate | scale down the weights of features for each boosting round | (0, 0.5) | 0.006 | 0.032 | 0.011 | 0.012 | 0.01 | 0.03 |
| reg_lambda | L2 regularization term on weights | {0, 0.01, 0.1, 1, 10} | 0.1 | 0 | 0.01 | 0.1 | 0.01 | 0.01 |
| n_estimators | total number of trees ensembled | {1000, 2000} | 2000 | 2000 | 2000 | 1000 | 1000 | 1000 |

## 4.4 Results and discussion

This section presents and discusses the experimental results to assess the performance of the proposed methods. We develop RNN-T and LSTM-T by adopting time embedding to RNN and LSTM, respectively. For consistency, we refer to the Informer with time embedding as Informer-T, while its counterpart without time embedding is labelled as Informer. Results of these DL methods, along with time embedding XGBoost, are presented and discussed, considering various factors such as the length of the input sequence, holiday traffic, data granularity, and the size of the training data. Additionally, the computational costs of the studied methods are discussed.

### 4.4.1 Forecasting capacity

We evaluate the forecasting capacity of evaluated methods as the forecasting horizon incrementally extends from 1 hour ($T_p = 1$) to 30 days ($T_p = 720$). The corresponding results in GEH are summarized in Figure 8.

We first discuss the effectiveness of different memory mechanisms adopted by RNN-based methods and Informer. It can be observed that when the prediction span extends to 7 days ($T_p = 168$) and beyond, LSTM generally outperforms RNN. This shows that forecasting a long-term (such as a week) imposes high demands on the model's ability to capture the long-range dependency from input sequences, which is achieved better with LSTM due to the effectiveness of its gated/cell state mechanisms in preserving useful information from a longer distance. Comparatively, RNN suffers from gradient vanishing, limiting its capability to learn information from an extended time ago. Meanwhile, for smaller forecasting horizons, the superiority of LSTM over RNN is less pronounced. This is likely because shorter-term traffic forecasting primarily revolves around capturing instantaneous variations, which is more based on recent past observations to account for traffic rapidly changing factors such as sudden events and real-time fluctuations in demand. Therefore, the requirement of remembering observations further back is low. In comparison to RNN-based methods, Informer consistently produces superior results across all forecasting horizons, including shorter periods like 1 hour or 6 hours. Moreover, its superiority becomes more pronounced as the forecasting period extends, with its performance showing a slower decline compared to RNN-based methods. Typically, a substantial performance gap can be observed for RNN-based methods when the prediction lengths are extended from 1 day (24 steps) to 7 days (168 steps), resulting in an average GEH increase of 75.36% when comparing LSTM to Informer on 7-day-ahead forecasting. The enhanced forecasting capability of Informer is related to its adoption of the Transformer framework, particularly the attention mechanism, which allows the model to assess the entire input sequence and effectively capture long-range dependency. Another significant factor is the generative decoding process employed by Informer, where the entire prediction sequence is generated simultaneously to prevent error accumulation. It is also worth noting that Informer maintained its superiority even in the case of short-term 1-step-ahead predictions because the attention mechanism allows it to scan and focus on the most relevant information in the input sequence, whether it be the most recent time steps for shorter-term forecasting or the long-range dependency required for longer-term forecasting.

The introduction of time embedding leads to a substantial improvement in the performance of RNN and LSTM, particularly in the context of long-term forecasting. After a minor increase in GEH when extending the forecasting span from 1 hour ($T_p = 1$) to 6 hours ($T_p = 6$) ahead, the performance of RNN-T and LSTM-T does not show a noticeable deterioration as the forecasting horizon further extends. In addition, despite their distinct abilities in retaining and passing historical information, there is only a slight difference in the accuracy of RNN-T and LSTM-T. This indicates that the prediction is made primarily based on current embedded time features. Moreover, predicting solely based on time features without access to recent traffic data, XGBoost demonstrates competitive results, especially for long-term forecasting. For instance, it delivers the best average results on both signalized arterial and freeway datasets for 30-day-ahead forecasting. Informer-T can sometimes enhance the performance of Informer for longer-term forecasting, such as 7 days or beyond, by aiding in learning periodicity patterns. However, neither Informer nor Informer-T can surpass RNN-based methods or XGBoost for such extended forecasting periods, potentially due to overfitting problems arising from their complexity.

We can therefore highlight that it seems that enhancing the ability to model periodicity patterns has a significant impact when the forecasting horizon reaches a certain size, which can be empirically determined as 7 days for the studied datasets. In contrast, for such extended forecasting periods, the inherent ability of the model to learn from the input sequence appears less important. In fact, for 30-day-ahead forecasting, RNN-T and XGBoost yield 31.1% and 34.04% lower GEH on average compared to Informer-T, respectively, and Wilcoxon rank-sum tests generated p-values of lower than 0.05 for both comparisons. Informer's inherently strong ability to extract information from input sequences seems more beneficial for forecasting relatively short forecasting spans (and in these instances, the use of time embedding may result in a decrease in its predictive accuracy). In these cases, the temporal dependencies learned from recent data sequences usually remain a factor with a certain impact, given their ability to offer more information about sudden changes in the traffic pattern.

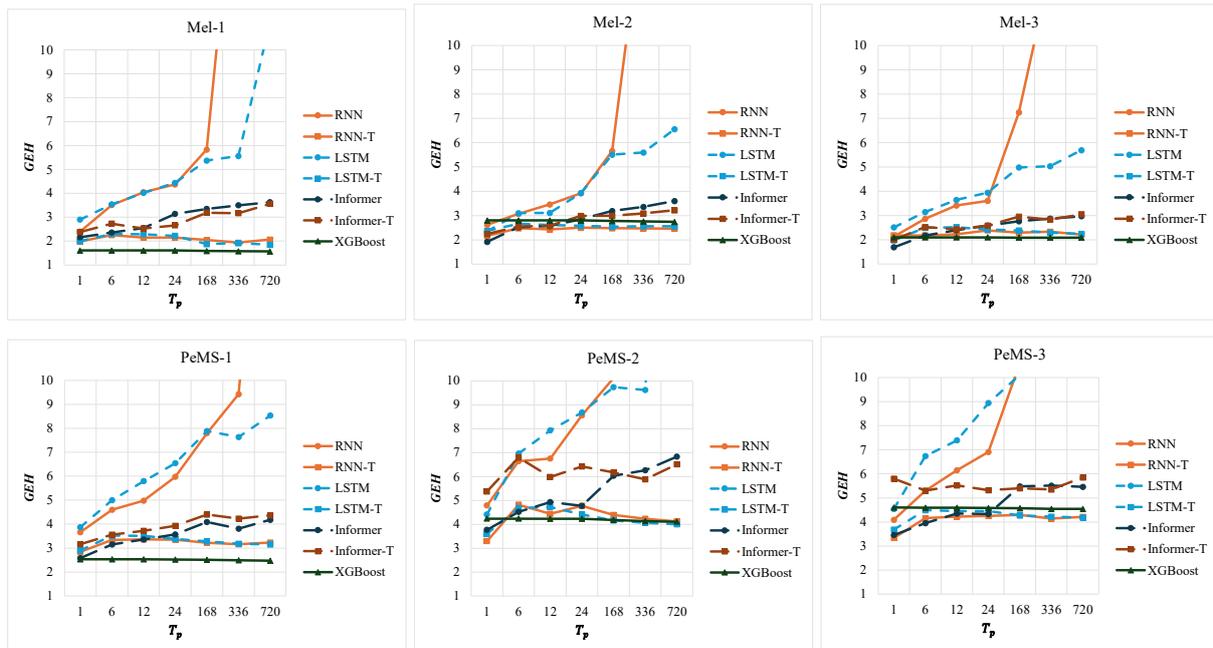

**Figure 8 Performance comparison of different methods with increasing forecasting span (for cases where the GEH exceeds 10, we limit the visualization to 10).**

We plot samples of visualized predicted flow for a single prediction in Figure 9. As depicted in Figure 9 (a), without time embedding, RNN tends to predict the traffic to be around the average value of the dataset after the first few steps, struggling to retrieve more information from the inputs. LSTM can capture a naive daily pattern (with traffic flow peaking near the morning peak) on the first day (24 steps). However, after that, it simply predicts the traffic to fluctuate around the mean value due to error accumulation and the limited ability to retain earlier useful information. Figure 9 (b) illustrates that the proposed refinement of time embedding has boosted the performance of RNN and LSTM, enabling the satisfactory capture of the daily pattern (a strong morning peak and a relatively subtle afternoon peak) and weekly pattern (the difference between weekdays and weekends), even to the last few steps of a 720-step prediction. The predictions of RNN-T and LSTM-T are also closely aligned with the output of XGBoost, indicating their significant reliance on time features. Figure 9 (c) shows that without time embedding, Informer can still generate various patterns similar to Informer-T, showcasing its strong forecasting capacity. However, their prediction is sub-optimal compared to simpler methods with time embedding, as seen in Figure 9 (b).

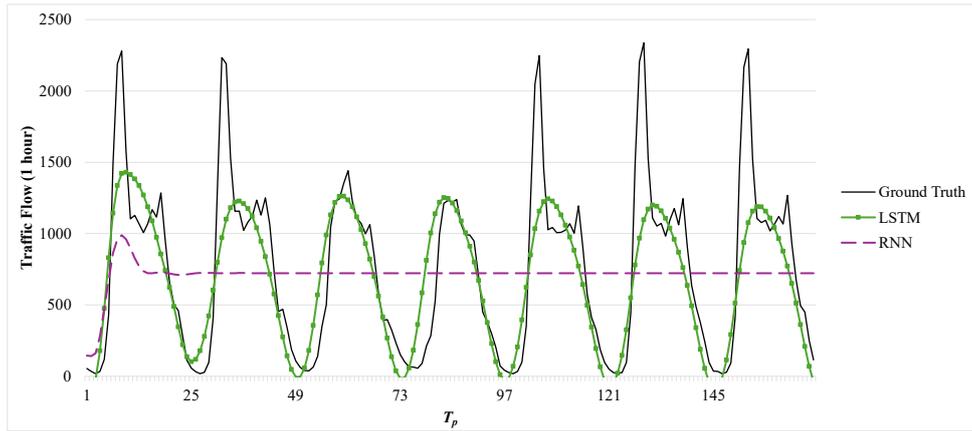

(a)

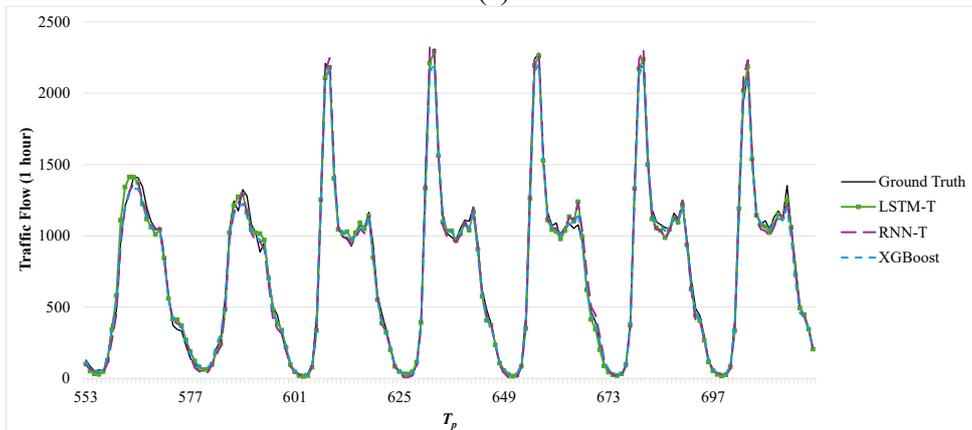

(b)

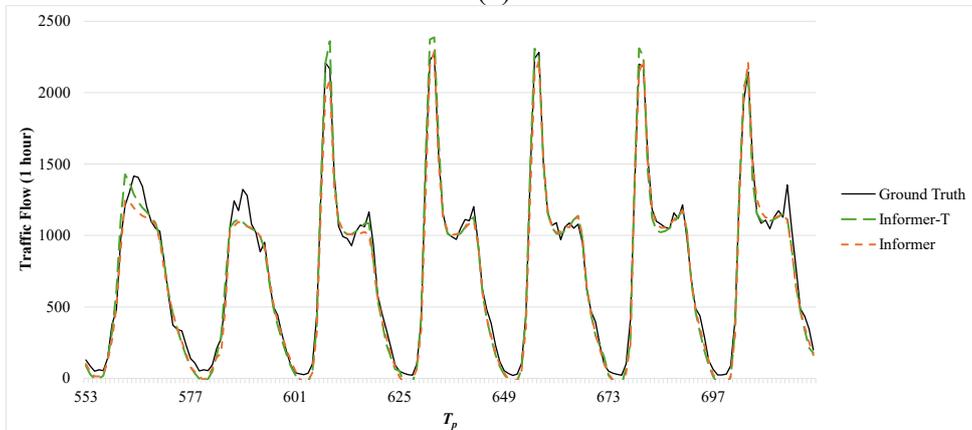

(c)

**Figure 9 Examples of visualized prediction for a single 720 steps prediction for Mel-1 from 0:00 August 1st to 23:45 August 30th, 2019, including (a): the first 168 steps prediction of RNN and LSTM, and the last 168 steps prediction of (b): RNN-T, LSTM-T and XGBoost and (c): Informer and Informer-T.**

To offer a better understanding of the models' performance, we present the prediction accuracy in widely used metrics for forecasting over the next 30 days (720 steps) in Table 2. To provide a more comprehensive comparison, we also include the results of common baselines, including SARIMA (Williams & Hoel, 2003), Temporal Convolutional Network (TCN) (S. Bai, Kolter, & Koltun, 2018), and Bidirectional LSTM (Bi-LSTM) (Abduljabbar, Dia, & Tsai, 2021). From the outcomes of this challenging long-term forecasting task, it can be observed that, without time embedding, the RNN-based

Page 17 of 32

methods (RNN, LSTM, and BiLSTM) do not exhibit a clear advantage over SARIMA, while Informer demonstrates a clear superiority. When incorporating time embedding, all of our studied methods greatly outperform SARIMA and TCN, with the worst-performing method (Informer-T) delivering 50.18% lower average GEH over the best baseline (TCN). It is also worth noting that, although Bi-LSTM practically doubles the size of the hidden state and a hidden memory cell state in LSTM, it does not lead to any evident accuracy improvement, suggesting that increasing the model size alone might not be an effective measurement of handling long-term prediction.

Table 2 Results of accuracy comparison for 30 days ahead forecasting

| Dataset No. | Metric | SARIMA[a] | TCN[b] | BiLSTM[c] | RNN | RNN-T | LSTM | LSTM-T | Informer | Informer-T | XGBoost |
|---|---|---|---|---|---|---|---|---|---|---|---|
| **Mel-1** | MAE | 244.6 | 161.99 | 486.77 | 498 | 46.54 | 280.75 | 46.47 | 75.05 | 69.63 | 40.2 |
| | RMSE | 422.23 | 220.34 | 573.57 | 581.66 | 70.43 | 419.5 | 74.6 | 131.77 | 115.76 | 62.25 |
| | $MAPE_{100}$ | 40.17% | 31.20% | 85.06% | 87.00% | 8.11% | 47.26% | 8.04% | 13.73% | 12.93% | 6.26% |
| | GEH | 9.21 | 7.43 | 18.94 | 19.37 | 2.07 | 11.19 | 1.85 | 3.63 | 3.56 | 1.57 |
| **Mel-2** | MAE | 162.38 | 127.61 | 156 | 374.17 | 57.85 | 157.25 | 61.41 | 82.21 | 70.21 | 65.25 |
| | RMSE | 245.38 | 167.79 | 214.4 | 447.68 | 85.45 | 218.53 | 90.37 | 124.7 | 104.85 | 94.41 |
| | $MAPE_{100}$ | 31.80% | 28.75% | 29.50% | 80.33% | 11.14% | 30.15% | 11.68% | 15.76% | 13.94% | 11.95% |
| | GEH | 6.59 | 6.37 | 6.46 | 16.02 | 2.46 | 6.56 | 2.56 | 3.6 | 3.22 | 2.74 |
| **Mel-3** | MAE | 158.55 | 121.06 | 180.82 | 308.2 | 53.15 | 140.18 | 53.5 | 67.85 | 68.13 | 49.97 |
| | RMSE | 250.89 | 160.25 | 279.78 | 389.2 | 83.4 | 211.27 | 82.13 | 110.34 | 107.78 | 74.39 |
| | $MAPE_{100}$ | 29.96% | 28.49% | 30.44% | 76.71% | 9.64% | 28.06% | 10.01% | 13.39% | 13.19% | 9.09% |
| | GEH | 6.44 | 5.85 | 7.12 | 13.12 | 2.21 | 5.69 | 2.24 | 2.97 | 3.05 | 2.09 |
| **PeMS-1** | MAE | 366.63 | 328.97 | 422.07 | 1006.95 | 130.13 | 357.88 | 132.7 | 156.47 | 162.58 | 103.72 |
| | RMSE | 531.26 | 429.6 | 557.29 | 1142.39 | 198.09 | 511.82 | 204.08 | 261.51 | 251.31 | 164.62 |
| | $MAPE_{100}$ | 24.61% | 31.51% | 36.66% | 161.48% | 9.73% | 27.71% | 9.16% | 13.68% | 14.13% | 7.35% |
| | GEH | 8.34 | 8.55 | 10.01 | 23.57 | 3.23 | 8.53 | 3.15 | 4.18 | 4.37 | 2.47 |
| **PeMS-2** | MAE | 1589.87 | 682.1 | 655.97 | 712.26 | 235.71 | 1845.42 | 235.21 | 342.16 | 323.45 | 231.81 |
| | RMSE | 28122.77 | 881.1 | 885.78 | 949.55 | 374.58 | 2205.03 | 388.86 | 512.23 | 480.99 | 343.1 |
| | $MAPE_{100}$ | 88.97% | 39.82% | 40.50% | 44.32% | 9.16% | 169.88% | 8.61% | 16.93% | 16.32% | 10.60% |
| | GEH | 12.85 | 13.28 | 12.4 | 13.33 | 4.12 | 31.62 | 4 | 6.83 | 6.51 | 4.11 |
| **PeMS-3** | MAE | 622.27 | 581.37 | 631.82 | 1792.41 | 220.67 | 627.8 | 226.04 | 268.78 | 275.83 | 242.67 |
| | RMSE | 912.39 | 752.98 | 880.72 | 2005.82 | 337.11 | 875.99 | 340.78 | 423.11 | 422.17 | 338.08 |
| | $MAPE_{100}$ | 24.24% | 31.79% | 27.20% | 144.01% | 9.67% | 26.20% | 9.38% | 13.18% | 14.50% | 10.79% |
| | GEH | 10.92 | 11.83 | 11.4 | 32.41 | 4.21 | 11.23 | 4.18 | 5.46 | 5.85 | 4.54 |

[a] The optimal order of SARIMA was determined with the Akaike Information Criterion and a period for seasonal differencing of 24.

[b] For TCN, we adopted the architecture in (S. Bai et al., 2018) and we designed 6 residual blocks with the number of filters set as 512 and the filter size set as 15. A linear layer was applied to the last step of the output to obtain predictions, and the training configurations (input length, batch size, loss function, initial learning rate, learning rate decay, optimizer, and early stopping strategy) were maintained in accordance with studied DL methods. Moreover, same time features were embedded into the input of TCN with time embedding.

[c] Bi-LSTM contains two isolated LSTMs to process data sequences in both forward and backward directions in the encoder, and the resulting hidden states/hidden memory cell states are concatenated. We adopt the same multilayer Seq2Seq framework and implementation details for BiLSTM as we did for the studied RNN-based methods.

### 4.4.2 Impact of input length

In this section, we investigate the impact of input length on forecasting accuracy. A long input sequence can offer the model a diverse set of traffic patterns and make it possible to capture long-range dependency. However, effectively learning from excessively long sequences is challenging due to problems like gradient vanishing and computational resource constraints. Figure 10 demonstrates the chances of performance as the length of input sequences varies.

Without time embedding, Informer yields better results than LSTM across different input lengths, indicating its superiority still exists even with short input sequences like 6 steps. Meanwhile, for both methods, inputting 6 hours of data is insufficient even for 1-hour-ahead forecasting, and as the forecasting horizon extends, even inputting 12 and 24 hours typically leads to relatively inferior outcomes compared to longer sequences. This shows the necessity of considering long-range dependency for long-term forecasting. When the forecasting horizon reaches 168 steps (7 days), the performance difference between $T = 168$ (7 days) and $T = 24$ (1 day) becomes evident for Informer. This is likely because forecasting over 7 days requires at least a week of data to model the strong weekly patterns in the traffic data. Note that we still acknowledge that the model's performance could be improved with input sequences longer than a week, as it can convey richer information like the monthly patterns observed in Figure 7, and we adopt $T = 720$ in all other tests, as it generates the best average performance for both LSTM and Informer. When time embedding is adopted, all input lengths produce very close results for both LSTM-T and Informer-T, indicating that forecasting based on time features can reduce the requirements for the length of the input sequence and the model's ability to learn from it. It is also worth noting that with a sufficiently long input sequence ($T \geq 168$), the performance of Informer is close to Informer-T, showing its strong learning ability seems enough to capture adequate week-level periodicity patterns, which LSTM fails to achieve even when the impact of error accumulation has been removed (for 1-step-ahead forecasting). Moreover, as the input length further increases to 1440 steps, a performance deterioration can be observed in some cases, especially for Informer/Informer-T, which might be due to overfitting the temporal noises when given an excessively long input sequence.

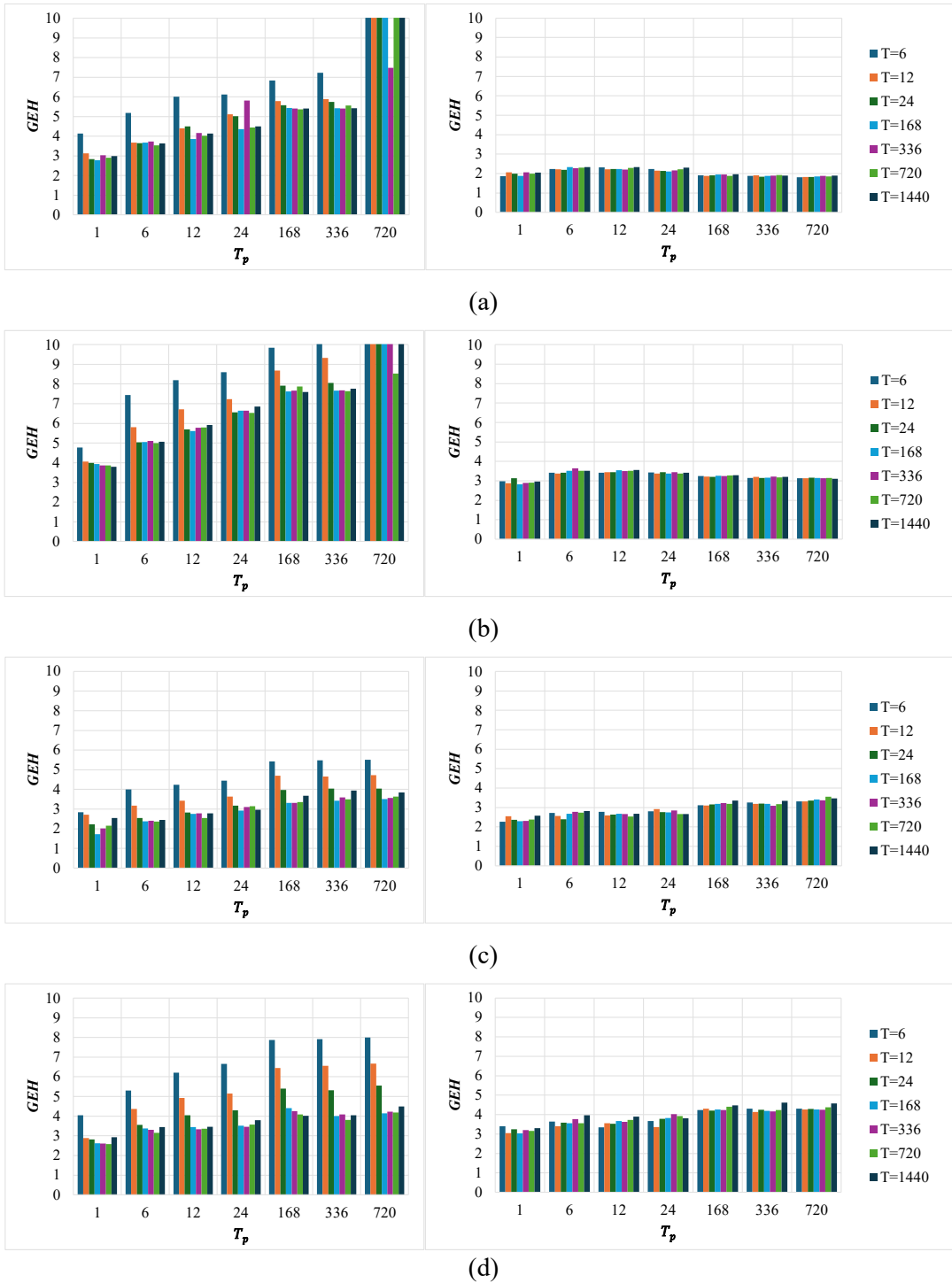

**Figure 10** Impact of different input lengths for (a): LSTM (left) and LSTM-T (right) on Mel-1, (b): LSTM (left) and LSTM-T (right) on PeMS-1, (c): Informer (left) and Informer-T (right) on Mel-1, and (d): Informer (left) and Informer-T (right) on PeMS-1. T represents the length of the input sequence. For some cases where the GEH exceeds 10, we limit the visualization to 10.

### 4.4.3 Impact of holiday traffic

Compared to prominent seasonal behaviours of traffic, such as daily and weekly patterns, event factors can impact traffic in a more stochastic way (Lana et al., 2018). Examples of events include traffic incidents, holidays, extreme weather, roadworks, etc. They normally occur at a non-fixed frequency and may coincide with daily/weekly traffic patterns. Here we investigate the impact of holiday traffic as a representative example. As mentioned before, in time embedding, we incorporate the holiday information by representing holiday with 1 and non-holiday with 0. Note that while holidays and events like roadworks are planned, other recurring events being difficult to predict can normally be assumed for what-if analysis, making them also known information and incorporated into prediction models in a similar way.

To investigate the impact of holidays on traffic flow forecasting, we evaluate the forecasting results for holiday and non-holiday week-long periods for Mel-1 and PeMS-1. For Mel-1, we select the holiday period from 25th to 31st December, where the 25th and 26th are public holidays. For PeMS-1, we study the holiday period from the 4th to the 10th of July, where the 4th of July is a public holiday. We also collect the results from a typical non-holiday week starting from 1st August. As shown in Figure 11, in long-term forecasting settings (predicting 720 steps), all methods exhibit significantly lower accuracy when predicting a period with holiday traffic (Holiday-720) compared to their performance in predicting an equivalent duration of non-holiday traffic (Non-Holiday). Specifically, there is a corresponding GEH increase of 173.51% and 61.24% for Mel-1 and PeMS-1, respectively, averaging across the four studied methods. A comparison between Informer and Informer-T indicates that time embedding can lead to an accuracy improvement in forecasting holiday traffic. Moreover, while this shows that time embedding can help with holiday forecasting to a certain extent, it cannot bring the prediction accuracy for holiday and non-holiday periods to the same level for long-term forecasting. Short-term 1-step ahead forecasting for holiday traffic is much less challenging for DL methods, especially those with a better ability to learn from input sequence (Informer/Informer-T), as they can adapt to recent pattern shifts.

We further explore the effect of time-embedding on holiday traffic forecasting with the prediction depicted in Figure 12 and local interpretations generated from Shapley Additive exPlanations (SHAP) (Lundberg & Lee, 2017). Figure 12 demonstrates that on public holidays, 25th and 26th December, there are abnormal traffic patterns compared to typical weekly patterns. Time embedding helps LSTM-T, Informer-T, and XGBoost to capture this pattern change, while Informer can hardly react to it. However, there also exist instances not marked as holidays but still exhibit similar abnormal behaviours (27th, 30th, 31st December). The pattern of these instances cannot be learned directly from holiday features and can cause considerable errors in some cases. Nonetheless, we observe that some of them can be accurately predicted with a deep understanding of high-level seasonality from time embedding. For instance, XGBoost reacts to the pattern change for these three days, demonstrating the potential of time embedding. The mismatch between the provided (pre-planned) holiday information and all the cases with abnormal holiday traffic patterns may stem from phenomena like it may be popular for local transport participants to take their annual leaves during the period from 27th to 31st December.

We apply SHAP to XGBoost to interpret the observations discussed earlier. SHAP is a widely used ML interpretation technique for estimating the contributions of each feature (Mangalathu, Hwang, & Jeon, 2020; Parsa, Movahedi, Taghipour, Derrible, & Mohammadian, 2020). The interpretations for the data points at 8:00 on 26th and 27th December are shown in Figure 13. It can be observed that while marking as a public holiday can greatly reduce the predicted traffic flow, for a day marked as a non-public holiday (27th December), XGBoost learned that the last few days in a year (a large number of *day of the year*) can heavily reduce the prediction, both leading to the accurate forecasting results shown in Figure 12. These effects become essential complements to the widely used *hour of the day* and *day of the week* features representing the dominant daily and weekly patterns.

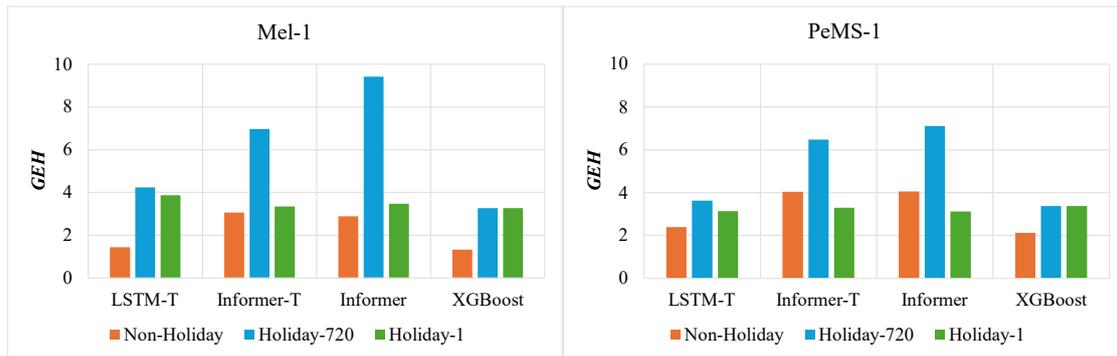

**Figure 11 Performance comparison when the accuracy calculation considers various 7-day periods:** Non-Holiday refers to the results for the first 7 days of a single 720 steps forecasting starting on 1st August (a typical week without any holidays), Holiday-720 for Mel-1 and PeMS-1 refers to the results of the studied holiday periods occurring in the last and first 7 days of a single 720-step forecasting, respectively, and Holiday-1 denotes the results for 1-step-ahead forecasting for the studied holiday periods.

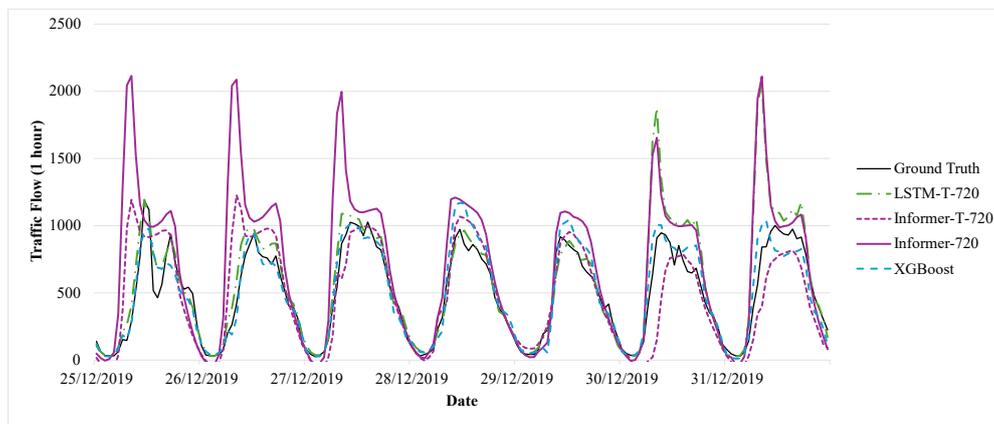

**Figure 12 Visualized prediction for a single 720-step prediction for the studied holiday traffic period for Mel-1.**

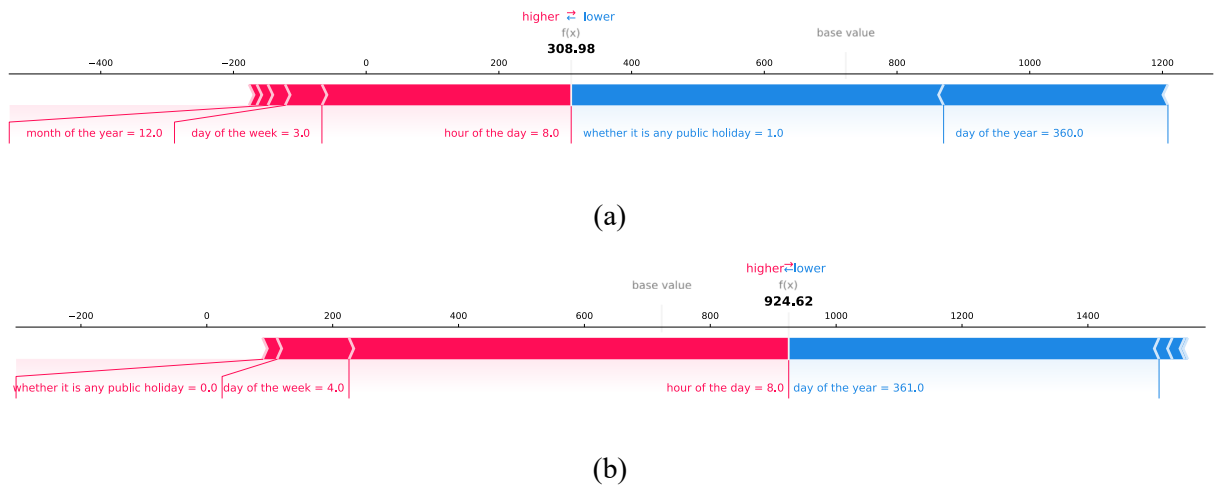

(a)

(b)

**Figure 13** Interpretation of the prediction from XGBoost for two instances in a single 720-step prediction for Mel-1, including (a): the instance of 8:00 December 26[th] and (b): the instance of 8:00 December 27[th]. Interpretations are generated by SHAP, which specifies an additive feature attribution method to show how the features of an instance contribute to the prediction. Red right-facing arrows indicate positive contributions (increasing prediction), while blue left-facing arrows indicate negative contributions (reducing prediction). Longer arrows represent greater impact.

### 4.4.4 Impact of data granularity

Different data granularity (acquired from different aggregation levels) is associated with different ITS applications, making it necessary to discuss the performance of predicting models when there is a change in the data granularity (Vlahogianni, Karlaftis, & Golias, 2014). In light of this, we additionally train and evaluate LSTM, LSTM-T, Informer, Informer-T, and XGBoost with 15-minute interval data. To facilitate a meaningful comparison with the results corresponding to the 1-hour interval data we have studied so far, we keep the actual forecasting horizons consistent. Specifically, when evaluating the methods with 15-minute interval data for 1 hour, 1 day, 7 days, and 30 days forecasting, the prediction length is scaled to 4, 96, 672, and 2880 steps, respectively. We set the input length to 2880, ensuring a consistent supply of 30 days' worth of data as input. Since the output from models are 15-minute level predictions, 4 neighbouring results are summed up and considered hourly traffic for GEH calculation, making the results corresponding to 1-hour and 15-minute granularity data comparable.

We present the results in Figure 14. It can be observed that training with 15-minute interval data can improve the accuracy of LSTM for relatively small forecasting horizons like 1 hour and 1 day ahead, which is likely attributed to the higher resolution data in the input sequence providing more information about local pattern changes. As the forecasting horizon extends to 7 days, the performance of LSTM and LSTM-15 become similar, which is in line with our finding that the focus of traffic forecasting gradually transitions from learning recent observations to periodicity modelling. With time embedding, the results of LSTM-T and XGBoost do not change significantly with higher data granularity, and they still generate the optimal performance when forecasting one day and beyond. Note that this does not necessarily mean that traffic data with smaller intervals like 15 minutes cannot provide additional meaningful information to periodicity learning. Instead, it may result from the combined influence of richer periodicity information and noise. The larger portion of the noise of the traffic time series of higher data resolution can also exacerbate the overfitting issue of more complex models (Informer/Informer-T), leading to observed accuracy decay compared with their counterparts trained with 1-hour interval data (average GEH increase of 9.88%).

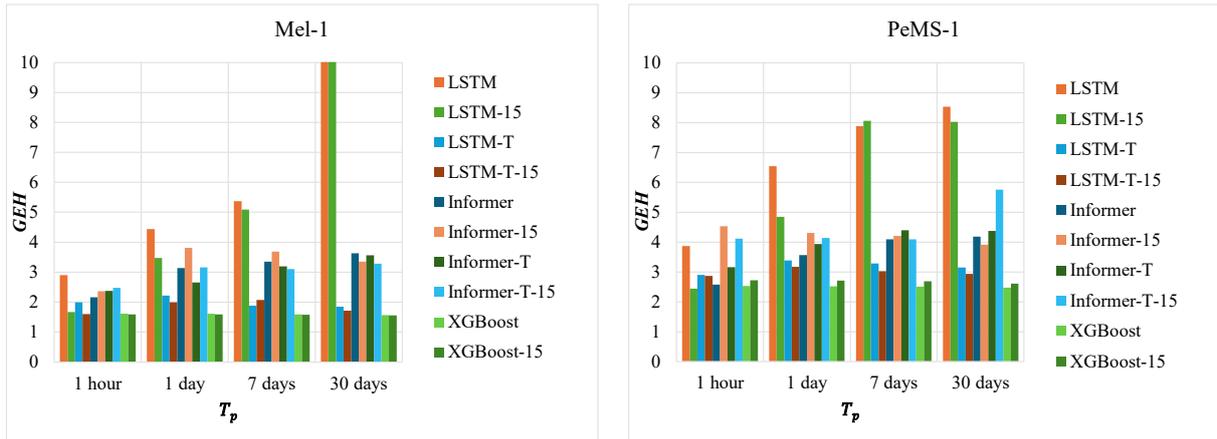

**Figure 14** Performance comparison considering a higher data granularity (for cases where the GEH exceeds 10, we limit the visualization to 10). LSTM-T-15, Informer-15, Informer-T-15, and XGBoost-15 refer to the case where LSTM-T, Informer, Informer-T, and XGBoost are trained and evaluated with 15-minute interval data, respectively.

### 4.4.5 Impact of the size of training data

A common challenge in practical scenarios is that the amount of available data is often limited. An ideal method should exhibit stable accuracy even when facing constraints on the quantity of available data. Therefore, we further assess the robustness of representative methods on 30-day forecasting when gradually reducing the amount of training data while fixing the validation and test set.

As shown in Figure 15, for DL methods, there is a performance decay when training data is reduced from 20 months to 12 months, with a more pronounced increase in GEH observed when the training data is further truncated to six months. This underscores the importance of including training data that spans an extended duration, ensuring a more substantial pool of training instances while maintaining reasonable data granularity. Furthermore, the sequential training setup (using input data sequence to predict subsequent values) of DL methods renders them less suitable in scenarios with considerably limited data collection periods. In our case where both input and output span 30 days, DL methods are not applicable when only two months of training data are available. In comparison, XGBoost demonstrates strong performance even with only one month of training data.

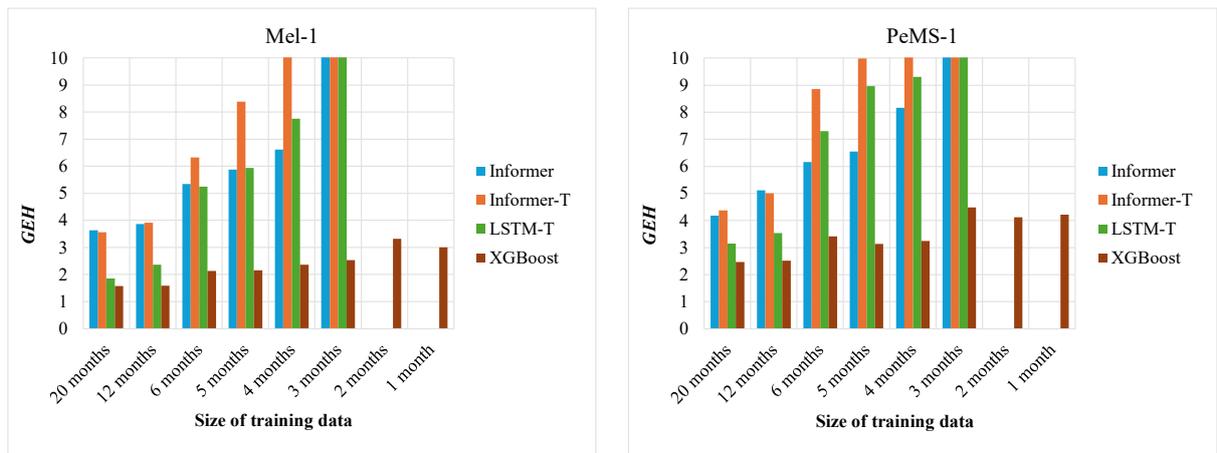

**Figure 15** The change of performance with reduced size of training data for 720-step prediction (for cases where the GEH exceeds 10, we limit the visualization to 10).

### 4.4.6 Computation cost

To provide a comprehensive evaluation, we examine the required computation time for the studied methods. The typical training time and inference time on the dataset Mel-1 for forecasting 30-day-ahead are analysed. We report the total time for the computation on the test set as inference time. The results for DL methods are summarised in Figure 16, which indicates that LSTM is slightly more time-consuming than RNN, while Informer exhibited the best efficiency. Seq2Seq RNN-based methods process data sequentially, which inherently limits parallelisation and impacts computational efficiency, especially when handling long input and/or output sequences. Based on the Transformer architecture, Informer enables better parallel processing. Moreover, its *ProbSparse* attention mechanism and generative-style decoder effectively reduce computational costs. In addition, while considerably enhancing the model's performance, the typical training time when time embedding is adopted increased by no more than 4 minutes per epoch, with the increase in inference time being less than 1 minute.

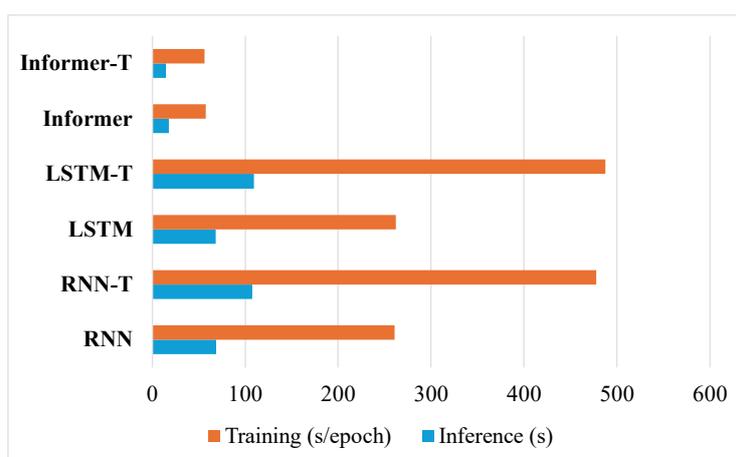

**Figure 16 The typical computation time for DL methods.**

The efficiency of XGBoost is remarkably higher than DL methods. The total training time for XGBoost is 3.08 seconds, and the inference time for the whole year prediction is merely 0.32 seconds. Our proposed time-embedding XGBoost leverages its high computational efficiency and scalability for tabular datasets, benefiting from optimised parallel processing and a tree-based architecture.

Furthermore, it is worth noting that the quantitative computation cost analysis is based on designated hardware, specifically, an advanced GPU (NVIDIA A100 GPU with 80GB RAM). While this hardware configuration is sufficient for real-time forecasting during inference stage, it indicates that DL methods require specialised hardware not only to handle extended training times but also to manage substantial memory usage (particularly for Informer, as the attention mechanism, even in its sparse form, demands significant memory). As a result, DL methods are resource-intensive, limiting their scalability for large-scale applications (e.g., training on larger datasets or deployment for many road segments). This necessitates trade-offs, such as adjusting the length of the historical input sequence or tuning the level of sparse attention to balance efficiency and performance. On the other hand, time-embedding XGBoost requires fewer computational resources, with lower memory overhead and reduced GPU dependency compared to DL models, making large-scale implementation less challenging. However, this comes at the cost of not being able to directly model temporal dependencies, making feature engineering essential.

# 5  Summary and conclusion

Accurate traffic forecasting plays a crucial role in ITS, for which ML methods have shown great potential. While existing research primarily focuses on short-term forecasting, there is a pressing need for longer-term predictions to cater to specific applications. In this study, a range of DL methods (Informer and multilayer Seq2Seq RNN and LSTM) and one ensemble ML method (XGBoost) were selected from the literature review as promising methods for long-term traffic prediction. In addition, we refined these methods by embedding seven types of time features into the input (time embedding). Developed methods were evaluated with large-scale real-world flow datasets collected from both signalized arterials and freeways, which is considered the first attempt to discuss their capability of predicting an excessively long horizon of up to 30 days. The impact of various important aspects of traffic forecasting was also assessed. Findings from this study include:

1. The introduced time embedding is indispensable for DL methods when the forecasting horizon extends to a certain level (which we empirically determined to be 7 days). This is because, for such a long forecasting horizon, the development of effective solutions should prioritize periodicity modelling over capturing temporal dependencies in sequential data. Consequently, time embedding greatly improves the performance of multilayer Seq2Seq RNN and LSTM to the same level, which otherwise could not generate satisfying accuracy, as they suffer from memory loss and error accumulation. With time embedding, these methods can outperform the state-of-the-art Transformer-based method, Informer-T, by 31.1% for 30-day-ahead forecasting, in the case of RNN as a comparison, while the latter seems to be impacted by the overfitting issues due to the complexity of the model. Moreover, predicting exclusively with time features, time embedding XGBoost delivers competitive results, including the best average accuracy on both signalized arterial and freeway datasets for 30-day-ahead forecasting.

2. The assessment of temporal correlation modelling capabilities among the studied DL methods indicates that Informer consistently outperforms LSTM and RNN, even in relatively short forecasting spans such as 1 hour or 6 hours. The performance gap becomes more pronounced as the forecasting horizon extends, reaching a substantial level for forecasting spans of 7 days and beyond. This suggests the potential of its attention mechanism/Transformer framework and generative style decoder. Moreover, Informer maintains its superiority for both long and short input sequences.

3. The analysis of the input length shows that a longer forecasting horizon normally demands a more extended input sequence and the capacity to effectively learn from it, and inputting less than 7 days (168 steps) can considerably affect the forecasting performance due to the strong week-level seasonality in the traffic data. Meanwhile, this requirement can be eased with time-embedding. While further increasing the input sequence length may offer potential benefits, it can also increase the risk of overfitting by accommodating more noise.

4. We studied the impact of holiday traffic as an attempt to capture the dynamics of event traffic. The results show that holidays can evidently impact long-term traffic forecasting, as the models struggle to adapt to recent pattern variations. Time embedding can help holiday forecasting by incorporating relevant features or learning high-level seasonality, and there is a notable impact stemming from the mismatch between pre-planned holiday dates and the actual time period with abnormal holiday traffic patterns.

5. Data with higher granularity (15-minute intervals) can provide more information on instantaneous pattern variations, helping improve forecasting accuracy, particularly for forecasting horizons up to 1 day. Moreover, it can provide more information for periodicity modelling and noise at the same time for time embedding. The impact of a larger portion of noisy data becomes more pronounced for rather complex models.

6. Analysis of the impact of limited training data reveals a significant decline in performance when six months or less of data are provided to DL methods. In contrast, XGBoost demonstrates decent results even with only one month of training data, indicating its superior robustness.

7. While resulting in a considerable performance improvement, time embedding only slightly increases computation time for DL methods (less than 4 minutes/epoch for training, and less than 1 minute for inference. In addition, XGBoost exhibits outstanding computation efficiency, with a training time of a few seconds for the large data set used in this experiment.

For future work, these findings can incentivize the research on better learning of seasonality and event-related patterns, as they are critical for achieving effective forecasting over longer horizons. Moreover, since XGBoost has demonstrated outstanding accuracy and robustness with very low computation cost, another line of work can be the improvement of its forecasting accuracy or interpretability. Furthermore, the data-driven methods can be further extended to the (macro) planning purposes by embedding planning datasets such as land use and demographics.